\documentclass[runningheads]{llncs}
\usepackage{graphicx}

\usepackage{tikz}
\usepackage{comment}
\usepackage{amsmath,amssymb} 
\usepackage{color}
\usepackage{blindtext}
\usepackage{ulem}
\usepackage[pagebackref,breaklinks,colorlinks,bookmarks=false]{hyperref}

\usepackage{graphicx}
\usepackage{float}
\usepackage{color}
\usepackage{colortbl}
\usepackage{multirow}
\usepackage{subfigure}
\usepackage{caption}

\let\llncssubparagraph\subparagraph
\let\subparagraph\paragraph
\usepackage[compact]{titlesec}
\let\subparagraph\llncssubparagraph

\def\onedot{.}
\def\eg{e.g\onedot} 
\def\ie{i.e\onedot}

\def\etal{et al\onedot}

\usepackage[accsupp]{axessibility}  

\usepackage[width=122mm,left=12mm,paperwidth=146mm,height=193mm,top=12mm,paperheight=217mm]{geometry}

\makeatletter
\newcommand{\printfnsymbol}[1]{%
        \textsuperscript{\@fnsymbol{#1}}%
}
\makeatother

\begin{document}
\pagestyle{headings}
\mainmatter

\title{DELTAR: Depth Estimation from a Light-weight ToF Sensor and RGB Image}

\titlerunning{DELTAR for Accurate Depth Estimation}

\author{
Yijin Li\inst{1} \and
Xinyang Liu\inst{1} \and
Wenqi Dong\inst{1} \and
Han Zhou\inst{1} \and
\newline
Hujun Bao\inst{1} \and
Guofeng Zhang\inst{1} \and
Yinda Zhang\inst{2}\printfnsymbol{1} \and
Zhaopeng Cui\inst{1}\thanks{Corresponding authors}
}

\authorrunning{Y. Li et al.}

\institute{
State Key Lab of CAD\&CG, Zhejiang University \and
Google
}

\maketitle

\begin{abstract}
Light-weight time-of-flight (ToF) depth sensors are small, cheap, low-energy and have been massively deployed on mobile devices for the purposes like autofocus, obstacle detection, etc. However, due to their specific measurements (depth distribution in a region instead of the depth value at a certain pixel) and extremely low resolution, they are insufficient for applications requiring high-fidelity depth such as 3D reconstruction. In this paper, we propose DELTAR, a novel method to empower light-weight ToF sensors with the capability of measuring high resolution and accurate depth by cooperating with a color image. As the core of DELTAR, a feature extractor customized for depth distribution and an attention-based neural architecture is proposed to fuse the information from the color and ToF domain efficiently. To evaluate our system in real-world scenarios, we design a data collection device and propose a new approach to calibrate the RGB camera and ToF sensor. Experiments show that our method produces more accurate depth than existing frameworks designed for depth completion and depth super-resolution and achieves on par performance with a commodity-level RGB-D sensor. Code and data are available at
\url{https://zju3dv.github.io/deltar/}.

\end{abstract}

\section{Introduction}
\begin{figure}[!t]
\begin{center}
    \includegraphics[width=1.0\linewidth]{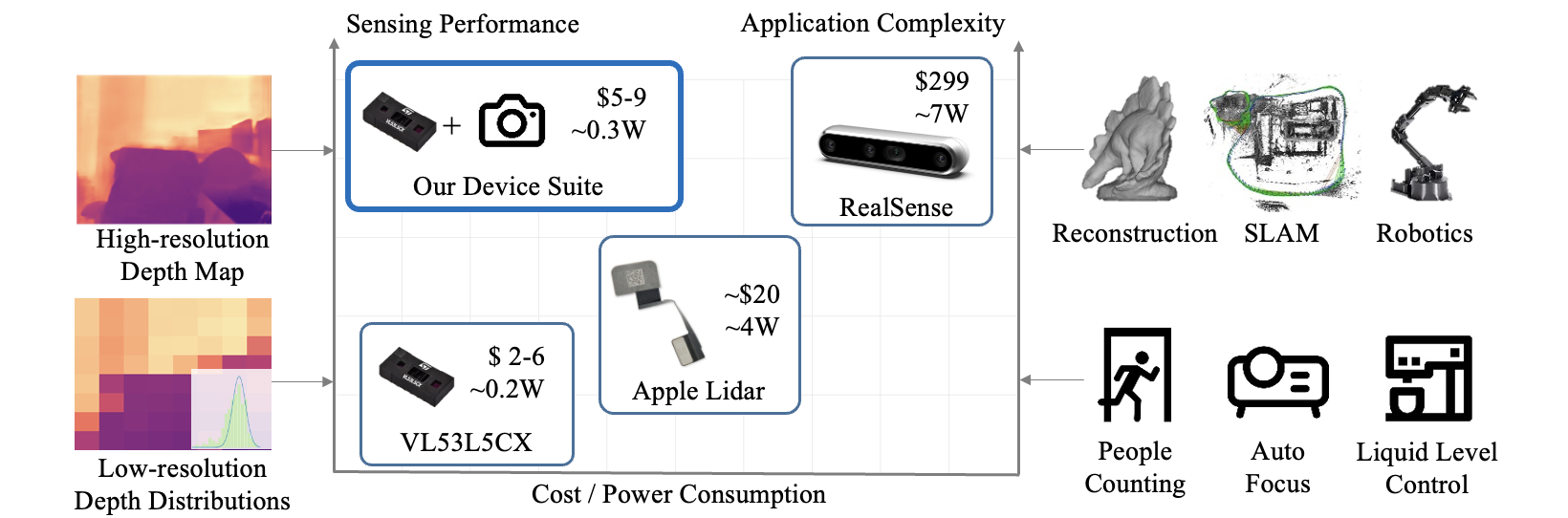}
\end{center}

\caption[]{Comparison between different depth sensors. Low-cost and low-power-consumed sensors like VL53L5CX are designed for simple applications such as people counting and autofocus. In this paper, we show how to improve the depth quality to be on par with a commodity-level RGB-D sensor by our DELTAR algorithm.\footnotemark[1]}
\label{fig:teaser}
\end{figure}

The depth sensor is a game changer in computer vision, especially with commodity-level products being widely available~\cite{kinect-fusion,dynamic-fusion,zhang-depth-completion,real-time-hand-tracking,bundle-fusion}.
As the main player, time-of-flight (ToF) sensors have competitive features
, \eg, compact and less sensitive to mechanical alignment and environmental lighting conditions.
and thus have become one of the most popular classes in the depth sensor market.
However, the price and power consumption, though already significantly lower than other technologies such as 
structured light (Microsoft Kinect V1), are still one to two orders of magnitudes higher than a typical RGB camera when reaching a similar resolution due to a large number of photons needs to be emitted, collected, and processed.
On the other hand, light-weight ToF sensors
are designed to be low-cost, small, and low-energy, which have been massively deployed on mobile devices for the purposes like autofocus, obstacle detection, etc
\cite{l5_sales}.
However, due to the light-weight electronic design, the depth measured by these sensors has more uncertainty (\ie, in a distribution instead of single depth value) and low spatial resolution (\eg, $< 10 \times 10$),
and thus cannot support applications like 3D reconstruction or SLAM~\cite{kinect-fusion,bundle-fusion}, that require high-fidelity depth
(see Fig.~\ref{fig:teaser}
).

\begin{figure}[!t]
\begin{center}
    \includegraphics[width=1.0\linewidth]{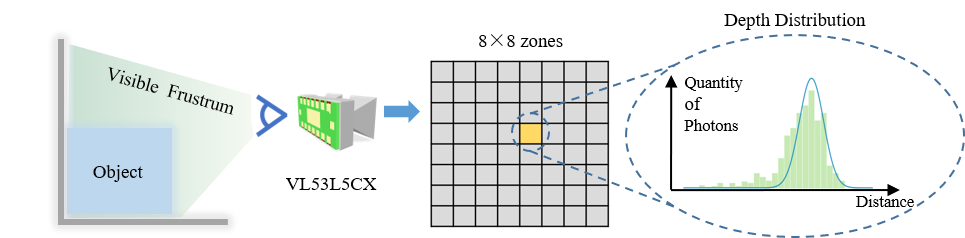}
\end{center}
\caption{L5 Sensing Principle. 
L5 has an extremely low resolution (8x8 zones) and provides depth distribution per zone.
} 
\label{fig:L5-principle}
\end{figure}

In contrast, RGB cameras are also widely deployed in modern devices with the advantage of capturing rich scene context at high resolution, but they are not able to estimate accurate depth with a single capture due to the inherent scale ambiguity of monocular vision.
We observe that these two sensors sufficiently complement each other and thus propose a new setting, i.e., estimating accurate dense depth maps from paired sparse depth distributions (by the light-weight ToF sensor) and RGB image.
The setting is essentially different from previous depth super-resolution and completion in terms of the input depth signal.
Specifically, the task of super-resolution targets relatively low-resolution consumer depth sensors, (\eg, $256\times192$ for the Apple LIDAR and $240\times180$ for the ToF sensor on the Huawei P30). In contrast, our task targets light-weight ToF sensors with several orders of magnitude lower resolution (\eg, $8\times8$), but provides a depth distribution per zone (see Fig.~\ref{fig:L5-principle}).
Depth completion, on the other hand, aims to densify incomplete dense high-resolution maps (\eg, given hundreds of depth samples), which is not available for light-weight ToF sensors.
Therefore, our task is unique and challenging due to the extremely low resolution of the input depth but accessibility to the rich depth distribution.

\footnotetext[1]{Icon credit: Iconfinder~\cite{iconfinder}}
To demonstrate, we use ST VL53L5CX
\cite{l5_page}
(denoted as L5) ToF sensor, 
which outputs $8\times8$ zones, each with a depth distribution, covering a total of 63$^{\circ}$ diagonal field-of-view (FoV)
and runs at a power consumption of about 200mW (vs. 4W of an Apple Lidar).
To fully exploit the L5 depth signals, we design DELTAR (\textbf{D}epth \textbf{E}stimation from \textbf{L}ight-weight \textbf{T}oF \textbf{A}nd \textbf{R}GB image), a neural network architecture tailored with respect to the underlying physics of the L5 sensors.
Specifically, we first build the depth hypothesis map sampled from the distribution reading of L5, and then use cross-domain attention to exchange the information between the RGB image and the depth hypothesis.
A self-attention is also run on image domain to exchange the information between regions covered by L5 and beyond, hence the output aligns with the RGB image and covers the whole FoV.
Experiments show that DELTAR outperforms existing architectures designed for depth completion and super-resolution, and improve the raw depth readings of L5 to maintain the quality on par with commodity-level depth sensors, such as Intel RealSense D435i.

Moreover, as no public datasets are available for this new task, we build a capturing system by mounting an L5 sensor and a RealSense RGB-D sensor on a frame-wire with reasonable field-of-view overlap.
To align the RGB image and L5's zones, we need to calibrate the sensors,
which is challenging as the correspondence cannot be trivially built between two domains.
To this end, we propose a new calibration method.
An EM-like algorithm is first designed to estimate the plane from L5 signals and then the extrinsic parameters between the L5 sensor and the color camera are  optimized by solving point-to-plane alignment in a natural scene with multiple planes.
With this capturing system, we create a dataset called ZJU-L5, which includes
about 1000 L5-image pairs from 15 real-world scenes with pixel-aligned RGB and ToF signals for training and evaluation purposes.
Besides the real-world data, we also simulate synthetic L5 signals using depth from NYU-Depth V2 dataset and use them to augment the training data.
The dataset is publicly available to facilitate and inspire further research in the community.

Our contributions can be summarized as follows. 
First, we demonstrate that light-weight ToF sensors designed for autofocus can be empowered for high-resolution and accurate depth estimation by cooperating with a color image.
Second, we prove the concept with a hardware setup and design a cross-domain calibration method to align RGB and low-resolution depth distributions, which enables us to collect large-scale data.
The dataset is released to motivate further research.
Third, we propose DELTAR,
a novel end-to-end transformer-based architecture, based on the sensors' underlying physics, can well utilize the captured depth distribution from the sensor and the color image for dense depth estimation.
Experiments show that DELTAR performs better than previous architectures designed for depth completion or super-resolution and achieves more accurate depth prediction results.

\section{Related Work}

\noindent\textbf{Monocular Depth Estimation.}
These methods predict a dense depth map for each pixel with a single RGB image.
Early approaches~\cite{saxenaLearningDepthSingle2005,ranftl2016dense,saxena2008make3d,shi2015break} use hand-crafted features or graphical models to estimate a depth map. More recent methods employ deep CNN~\cite{eigenDepthMapPrediction2014,7785097,xu2017multi,hao2018detail,xu2018structured,hu2019revisiting} due to its strong feature representation.
Among them,
some methods~\cite{bts,huynh2020guiding} exploit assumptions about indoor environments, \eg, plane constraints, to regularize the network.
Other methods~\cite{mixing_dataset,dpt} try to benefit from more large-scale and diverse data samples by designing loss functions and mixing strategies.
Besides, \cite{fu2018deep,adabins} propose to model depth estimation as a classification task or hybrid regression
to improve accuracy and generalization. 
Nonetheless, these methods cannot generalize well on different scenes due to their lack of metric scale innately.

\noindent\textbf{Depth Completion.}
Depth completion aims to recover a high-resolution depth map given some sparse depth samples and an RGB image.
Spatial propagation network(SPN) series methods~\cite{liuSPNLearningAffinity2017,chengCSPNDepthEstimation2018,chengCspnLearningContext2020,nlspn} are one of the most popular methods which learned local affinities to refine depth predictions. 
Recently, some works~\cite{chen2019learning,xu2019depth,qiu2019deeplidar} attempt to introduce 3D geometric cues in the depth completion task, e.g., by introducing surface normals as the intermediate representation, or learn a guided network~\cite{guidenet} to utilize the RGB image better.
More recently, PENet~\cite{penet} propose an elaborate two-branch framework, which reaches the state-of-the-art.
This type of method, however, is not suitable for our task because it assumes the pixel-wise depth-to-RGB alignment, while light-weight ToF sensors only provide a coarse depth distribution in each zone area without exact pixel-wise correspondence.

\noindent\textbf{Depth Super-Resolution.}
This task aims to boost the consumer depth sensor to a higher spatial resolution to match the resolution of RGB images.
Most early works are based on filtering~\cite{liu_super_reso_filter,yang_super_reso_filter} or formulate depth super-resolution as an optimization problem~\cite{park_super_reso_opt,diebel_super_reso_opt}. Later researches focus more on learning-based method~\cite{wang2020depth,prdepth,pnp-depth,hui-depth-super-reso}. 
Among them, 
Xia~\etal~\cite{prdepth} propose a task-agnostic network which can be used to process depth information from different sources. Wang~\etal~\cite{pnp-depth} iteratively updates the intermediate feature map to be consistent with the given low-resolution depth.
In contrast to these methods which usually take a depth map with more than 10 thousand pixels as input, our task targets light-weight ToF sensors with several orders of magnitude lower resolution (\eg, $8\times8$), but provides a depth distribution per region. 

\section{Hybrid Sensor Setup}

This paper aims to predict a high-resolution depth image from a light-weight ToF sensor (\eg, L5) guided by a color image. 
While no public datasets are available, we build a device with hybrid sensors and propose the calibration method for this novel setup.

\subsection{L5 Sensing Principle}

L5 is a light-weight ToF-based depth sensor. 
In conventional ToF sensors, the output is typically in a resolution higher than 
10 thousand pixels
and measures the per-pixel distance along the ray from the optical center to the observed surfaces.
In contrast, L5 provides multiple depth distributions with an extremely low resolution of $8\times8$ zones,
covering 63$^{\circ}$ diagonal FoV in total.
The distribution is originally measured by counting the number of photons returned in each discretized range of time, and then fitted with a Gaussian distribution (see Fig.~\ref{fig:L5-principle}) in order to reduce the broadband load and energy consumption since only mean and variance needs to be transmitted.
Due to the low resolution and high uncertainty of L5, it cannot be directly used for indoor dense depth estimation. Please refer to supplementary materials for more details about L5.

\begin{figure}[!t]
\centering
\subfigure[Device setup]{
\begin{minipage}{0.45\linewidth}
\centering
\includegraphics[width=0.9\linewidth]{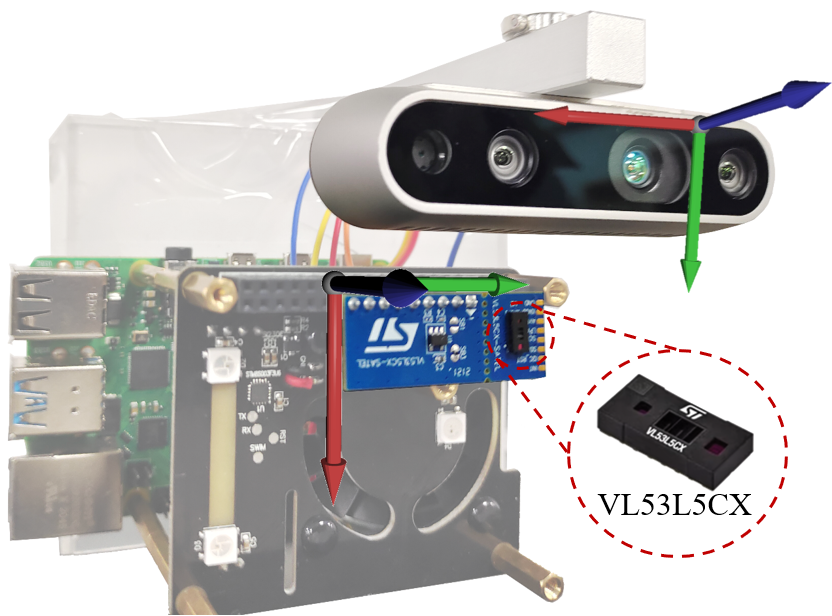}
\end{minipage}
}
\subfigure[Aligned L5's zones and color image]{
\begin{minipage}{0.45\linewidth}
\centering
\includegraphics[width=0.9\linewidth]{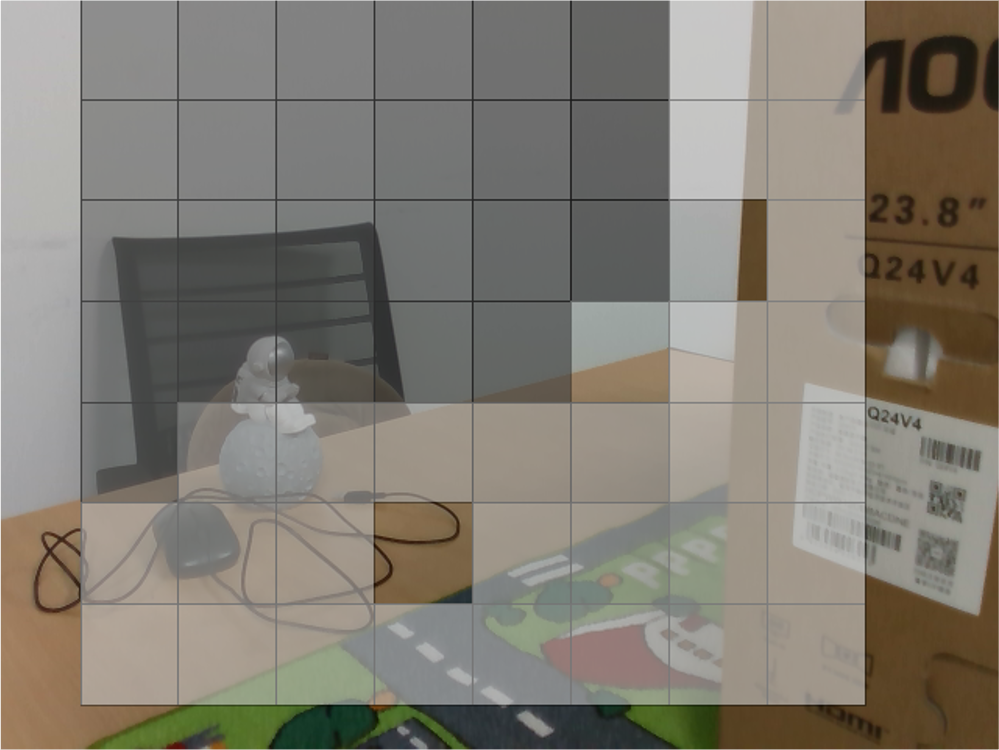}
\end{minipage}
}
\caption{Hybrid sensor setup. (a) We mount a L5 with an Intel RealSense 435i on a metal frame. (b) Blending color images with L5' depth. White color represents close range, black color represents long range. According to the valid status returned by L5, we hide all invalid zones which may receive too less photons or fail in measurement consistency.}
\label{fig:hybrid-setup}
\end{figure}

\subsection{Device Setup}

Fig.~\ref{fig:hybrid-setup}-(a) shows our proposed device suite.
An L5 and an Intel RealSense D435i are mounted on a metal frame facing in the same direction.
It is worth noting that we only used RealSense's color camera along with L5 when estimating depth, and the depth output by RealSense is used as the ground truth to measure the quality of our estimation. 
The horizontal and vertical FoV of the L5 are both 45$^{\circ}$, while the RealSense's color camera has 55$^{\circ}$ horizontal FoV and 43$^{\circ}$ vertical FoV. 
As a result,
the L5 sensor and the color camera share most of the FoV but not all in our setup.

\subsection{Calibration}

In order to align the L5 outputs with the color image, we need to calibrate the multi-sensor setup, \ie, computing the relative rotation and translation between the color camera and the L5 sensor.
Similar to the calibration between LIDAR and camera~\cite{lidar_calib_1}, we also calibrate our device suite by solving a point-to-plane fitting problem.
However, it is not trivial to fit a plane with raw L5 signals since it does not provide the pixel position of the depth value. We observe that, when facing a plane, in each zone $k\in Z$, there must be a location $(x_k, y_k)$, though unknown, whose depth is equal to the mean of the corresponding distribution $m_k$ returned by L5.
Therefore we can optimize both the plane parameter $\{n,d\}$ (the frame subscript is omitted for brevity) and the pixel position $(x_k, y_k)$ through:
\begin{equation} \label{eq:opt-plane}
\begin{aligned}
    \{n,d,x_k,y_k\mid k\in Z\} = &\mathop{\arg}\mathop{\min} \sum_{k\in Z} \| n\cdot K^{-1}(x_k,y_k,m_k)^T+d \|^2 \\
    &\textbf{s.t.}\ x_{\text{min}}^k\leq x_k \leq x_{\text{max}}^k,y_{\text{min}}^k\leq y_k \leq y_{\text{max}}^k,
\end{aligned}
\end{equation}
where $(x,y)_{(\text{min},\text{max})}^k$ is the boundary of the zone $k$ in L5 coordinates, and $K$ is the intrinsic matrix.
Clearly, Eq.~\ref{eq:opt-plane} is non-convex thus we solve it by an EM-like algorithm. Specifically, we first initialize all 2D positions at the center of the zone. In the E-step, we back-project these 2D points with measured mean depth, and then fit a 3D plane. In the M-step, we adjust the 2D positions within each zone by minimizing the distance of the 3D points to the plane.
The steps run iteratively until convergence. 
During the iteration, the points that are too far from the plane are discarded from the optimization.

We then obtain the extrinsic transformation matrix by solving a point-to-plane fitting problem.
We use our device suite to scan three planes that are not parallel to each other, and ensure that we only observe one plane most of the time.
We employ an RGB-D SLAM~\cite{orb-slam2}, which recovers from color images a set of camera poses and point cloud $P$
in real-world metric scale,
and each point belongs to a certain plane.

For each time stamp $i\in F$, we use $P_i$ to represent the subset of $P$ that are visible in frame $i$ and transformed from the world coordinate system to current RGB camera's. We also have the planar parameters $\{n_i \in \mathbb{R}^3,d_i \in \mathbb{R}\}$ (normal and offset to the origin) in the current L5's coordinate system by solving Eq.~\ref{eq:opt-plane}, then the extrinsic parameters can be solved by minimizing the point to plane distance:

\begin{equation} \label{eq:opt-extrinsic}
    \{R,t\} = \mathop{\arg}\mathop{\min} \sum_{i\in F}\sum_{p\in P_i} \| n_i\cdot (R\cdot p+t)+d_i \|^2,
\end{equation}
where $[R,t]$ are the transformation that map 3D points from the RGB camera's coordinate system to L5's.

For the device setup shown in Fig.~\ref{fig:hybrid-setup}-(a), we are able to recover the rotation transformation of the two sensors close to 90 degrees. The mean distances between the L5 measurement and the 3D point cloud before and after calibration are 7.5 cm and 1.5 cm, respectively. An example of aligned L5's zones and color image is shown in Fig.~\ref{fig:hybrid-setup}-(b).

\section{The DELTAR model}

\begin{figure}[!t]
\begin{center}
    \includegraphics[width=1.0\linewidth]{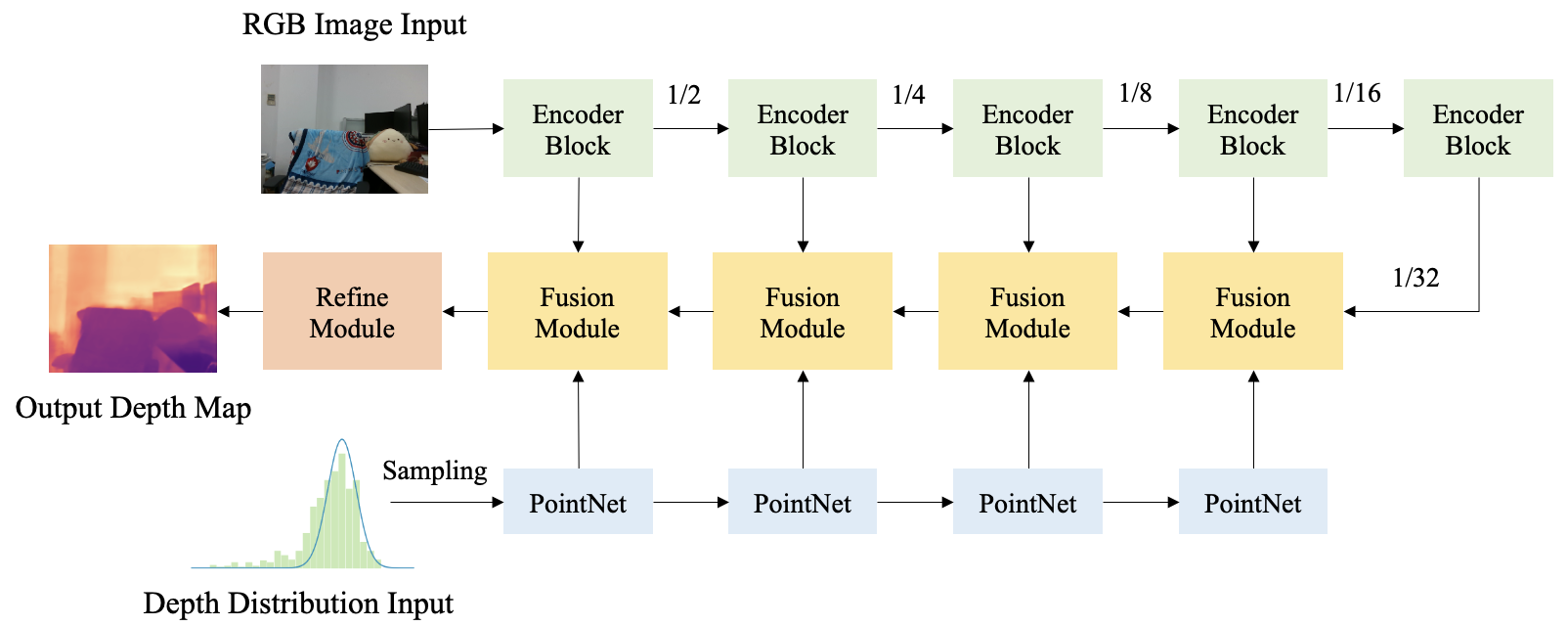}
\end{center}
\caption{Overview architecture of the fusion network. Our model takes depth distributions and color image as input, and fuse them at multiple-scale with attention based module before predicting the final depth map.}

\label{fig:pipeline}
\end{figure}

With the calibration, we are able to align the L5'zones with the color image. 
Based on the characteristic of each modality, we propose
a novel attention-based network to predict a high-resolution depth image given the aligned L5 signals and color image.
We first design a module to extract features from the distribution (Sec.~\ref{sec:feature-extraction}), and then propose a cross-domain Transformer-based module to fuse with the color image features at different resolutions (Sec.~\ref{sec:fusion-module}).
Finally, we predict the final depth values through a refinement module (Sec.~\ref{sec:refine-module}). 
An overview of the proposed method is shown in Fig.~\ref{fig:pipeline}.

\subsection{Hybrid Feature Extraction} \label{sec:feature-extraction}
Many works have been designed for fine-grained observations, such as RGB images, depth and point clouds. In contrast, how to extract features from distributions is barely studied.
A straightforward idea is to encode the mean and variance directly. However, the depth variance is often smaller than the mean by several magnitudes, which may make it difficult to train the network because of internal covariate shift~\cite{batch-norm}. 
In Section~\ref{sec:ablation} we show that directly encoding the mean and the variance does not work well in our experiments. Therefore, we propose to discretize the distribution by sampling depth hypotheses.
Instead of uniform sampling~\cite{patchmatch,sgm}, we uniformly sample on the inverse cumulative distribution function of the distribution, so the density of the sampling follows the distribution.
We utilize PointNet~\cite{pointnet} without T-Net
to extract features from the sampled depth hypotheses. 
Multiple pointnets are stacked to extract multi-level distribution features. 
We use a standard convolutional architecture for the color image, \ie, Efficient B5~\cite{efficient-net}, to extract multi-level features.
Unlike image feature extraction, we do not conduct a down-sample operation when distilling multi-level distribution features.

\begin{figure}[!t]
\begin{center}
    \includegraphics[width=1.0\linewidth]{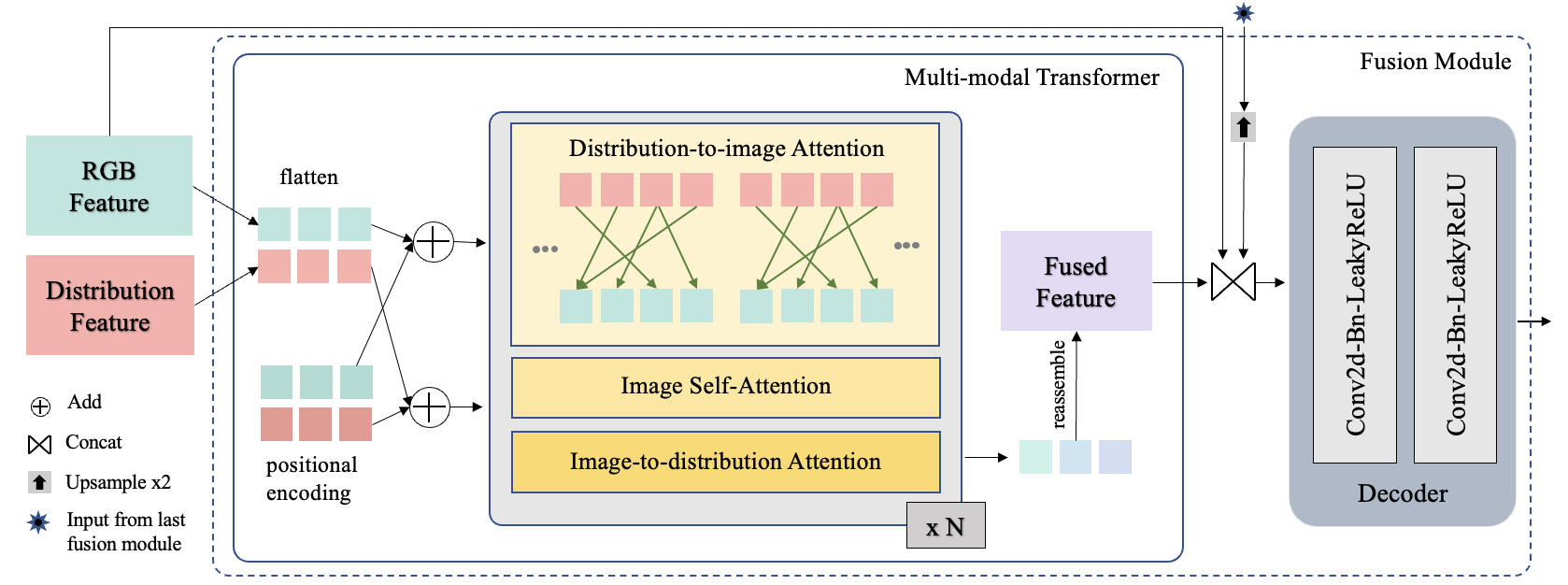}
\end{center}
\caption{Details of fusion module. The image and distribution features are flattened into 1-D vectors and added with the positional encoding. The added features are then fused by three different attention mechanisms and concatenated with last layer's feature and the skip-connected image features. Finally, a decoder decodes the concatenated feature and outputs it to the next fusion layer.} 
\label{fig:fusion-module}
\end{figure}

\subsection{Transformer-based Fusion Module} \label{sec:fusion-module}
The fusion module takes two multi-modal data, including image features and distribution features as input, and outputs fused features. 
In the task of depth completion and super-resolution, depth features and RGB features are usually concatenated or summed in the fusion step~\cite{qiu2019deeplidar,penet}. This may be sufficient for fine-grained observations which provide the pixel-wise depth-to-RGB alignment,  but it is not suitable for our task since pixel-wise alignment is not available between the depth hypothesis map and the RGB features. 
Inspired by the recent success of Transformer~\cite{transformer-survey,vit,flowformer}, we adopt attention mechanisms, which process the entire input all at once and learn to focus on sub-components of the cross-modal information and retrieve useful information from each other.

\noindent \textbf{Cross-attention Considering Patch-distribution Correspondence.}
The Transformer adopts an attention mechanism with the Query-Key-Value model. Similar to information retrieval, the query vector $Q$ retrieves information from the value vector $V$, according to the attention scores computed from the dot product of queries $Q$ and keys $K$ corresponding to each value.
The vanilla version of Transformer contains only self-attention, in which the key, query, and value vectors are from the same set. In multi-modal learning, researchers use cross-attention instead, in which the key and value vectors are from one modal data, and the query vectors are from the other.
We first conduct Distribution-to-image attention, that is, taking the key and value vectors from the distribution's feature, and the query vector from the image features, so that the network learns to retrieve information from the candidate depth space.
Considering that each distribution from L5 signals corresponds to a specific region in the image, we only conduct cross-attention between the corresponding patch image and the distribution (see Fig.~\ref{fig:hybrid-setup}-(b)).
In Sec.~\ref{sec:ablation}, we show that conducting cross-attention without considering the patch-distribution correspondence leads to severe performance degradation.
Empirically, we find that adding image-to-distribution attention leads to better performance.

\noindent \textbf{Propagation by Self-attention.}
It is not enough to conduct cross-attention as many regions on the image are not covered by the L5's FoV, and these regions cannot benefit from the distribution features. To propagate the depth information further, we also include image self-attention. This step helps the learned depth information propagate to a global context. Besides, the fused feature can be blended to make the feature map smoother.
We conduct cross-attention between the two modal data and self-attention over the image feature alternatively for $N$ times, as shown in Fig.~\ref{fig:fusion-module}. In our experiment, we set $N=2$.

\begin{figure}[!t]
\begin{center}
    \includegraphics[width=0.9\linewidth]{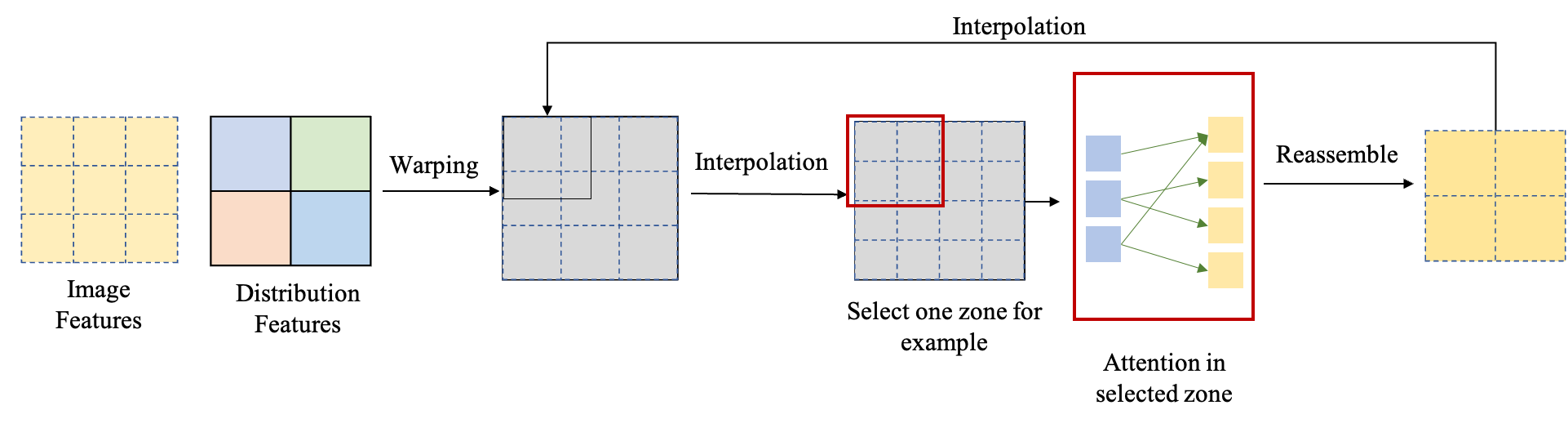}
\end{center}
\caption{Interpolation for solving misalignment. When the boundary of the L5's zones and the image feature cannot be aligned precisely, simply quantizing the floating-number boundary could introduce a large negative effect, so we propose to fuse feature after interpolation.}
\label{fig:interpolation}
\end{figure}

\noindent\textbf{Solving Misalignment by Interpolation.}
Misalignment occurs when warping L5 zones to an image. See Fig.~\ref{fig:interpolation} for a toy example.
Simply quantizing the floating-number boundary could introduce a largely negative effect, especially when the fusion is operated on low-resolution feature maps.
Moreover, the image resolution corresponding to each zone should be the same to facilitate putting them into a batch. To this end, we fuse on the interpolated feature and then interpolate the fused image features back.

\subsection{Refinement Module} \label{sec:refine-module}
We employ the mViT proposed in Adabins~\cite{adabins} as our refinement module to generate the final depth map. Unlike directly regressing depth, the refinement module predicts the depth as a linear combination of multiple depth bins.
Specifically, the refinement module predicts a bin-width vector $b$ per image and linear coefficient $l$ at each pixel. The depth-bin's centers $c(b)$ can be calculated from $b$. Suppose the depth range is divided into $N$ bins, the depth at pixel $k$ can be formulated as:

\begin{equation}
    d_k = \sum_{n=1}^{N}c(b_n)l_n.
\end{equation}

\subsection{Supervision}
Following~\cite{adabins,bts}, we use a scaled version of the Scale-Invariant loss (SI) introduced by~\cite{si-loss}:

\begin{equation} \label{eq:loss-function}
    \mathcal{L} = \alpha \sqrt{\frac{1}{T} \sum_i{g_i^2} - \frac{\lambda}{T^2} (\sum_i{g_i})^2},
\end{equation}
where $g_i=\log \tilde{d_i}-\log d_i$ defined by the estimated depth $\tilde{d_i}$ and  ground truth depth $\log d_i$,  and $T$ denotes the number of pixels with valid ground truth values. 
We use $\lambda=0.85$ and $\alpha=10$ for all our experiments.

\section{Experiment}
\subsection{Datasets and Evaluation Metrics}
\noindent \textbf{NYU-Depth V2 for Training.} We use the NYU-Depth V2 dataset to simulate and generate the training data containing L5 signals and color images,
from which we select a 24K subset following~\cite{bts,adabins}.
For each image, we select a set of zones and
according to the L5 sensing principle, we count the histograms of the ground true depth map in each zone and fit them with Gaussian distributions. The fitted mean and variance are used together with the color images as the input for network training. We exclude the depths beyond the L5 measurement range during the histogram statistics.

\noindent \textbf{ZJU-L5 dataset for Testing.}
Since the current datasets do not contain the L5 signals, we create an indoor depth dataset using the device suit in Fig.~\ref{fig:hybrid-setup}-(a) to evaluate our method. This dataset contains 1027 L5-image pairs from 15 scenes, of which the test set contains 527 pairs and the other 500 pairs are used for fine-tuning network. We show the results after fine-tuning in the supplementary material.

\noindent \textbf{Evaluation Metrics.}
We report the results in terms of standard metrics including 
thresholded accuracy ($\delta_i$),
mean absolute relative error (REL),
root mean square error (RMSE)
and average ($\text{log}_{10}$) error.
The detailed definitions of the metrics are provided in the supplementary material.

\begin{table}[!t]
    \centering

\tabcolsep 10pt
\begin{tabular}{lrrrrrr}
\hline
\multicolumn{7}{c}{\cellcolor[HTML]{D9EAD3}{\color[HTML]{2B2B2B} Comparison with Monocular Depth Estimation}} \\ \hline
Methods & \multicolumn{1}{l}{$\delta_1\uparrow$} & \multicolumn{1}{l}{$\delta_2\uparrow$} & \multicolumn{1}{l}{$\delta_3\uparrow$} & \multicolumn{1}{l}{REL$\downarrow$} & \multicolumn{1}{l}{RMSE$\downarrow$} & \multicolumn{1}{l}{$\text{log}_{10}\downarrow$} \\ \hline
VNL~\cite{vnl} & \multicolumn{1}{l}{0.661 } & \multicolumn{1}{l}{0.861 } & \multicolumn{1}{l}{0.928 } & \multicolumn{1}{l}{0.225 } & \multicolumn{1}{l}{0.653 } & \multicolumn{1}{l}{0.104 } \\ 
BTS~\cite{bts} & 0.739 & 0.914 & 0.964 & 0.174 & 0.523 & 0.079 \\ 
AdaBins~\cite{adabins} & 0.770 & 0.926 & 0.970 & 0.160 & 0.494 & 0.073 \\ 
Ours & \textbf{0.853} & \textbf{0.941} & \textbf{0.972} & \textbf{0.123} & \textbf{0.436} & \textbf{0.051} \\ \hline
\multicolumn{7}{c}{\cellcolor[HTML]{FFF2CC}{\color[HTML]{2B2B2B} Comparison with Depth Completion}} \\ \hline
Methods & \multicolumn{1}{l}{$\delta_1\uparrow$} & \multicolumn{1}{l}{$\delta_2\uparrow$} & \multicolumn{1}{l}{$\delta_3\uparrow$} & \multicolumn{1}{l}{REL$\downarrow$} & \multicolumn{1}{l}{RMSE$\downarrow$} & \multicolumn{1}{l}{$\text{log}_{10}\downarrow$} \\ \hline
PrDepth~\cite{prdepth} & 0.161 & 0.395 & 0.660 & 0.409 & 0.937 & 0.249 \\ 
NLSPN~\cite{nlspn} & 0.583 & 0.784 & 0.892 & 0.345 & 0.653 & 0.120 \\ 
PENet~\cite{penet} & 0.807 & 0.914 & 0.954 & 0.161 & 0.498 & 0.065 \\ 
Ours & \textbf{0.853} & \textbf{0.941} & \textbf{0.972} & \textbf{0.123} & \textbf{0.436} & \textbf{0.051} \\ \hline
\multicolumn{7}{c}{\cellcolor[HTML]{DFF8FF}{\color[HTML]{2B2B2B} Comparison with Depth Super Resolution}} \\ \hline
Methods & \multicolumn{1}{l}{$\delta_1\uparrow$} & \multicolumn{1}{l}{$\delta_2\uparrow$} & \multicolumn{1}{l}{$\delta_3\uparrow$} & \multicolumn{1}{l}{REL$\downarrow$} & \multicolumn{1}{l}{RMSE$\downarrow$} & \multicolumn{1}{l}{$\text{log}_{10}\downarrow$} \\ \hline
PnP-Depth~\cite{pnp-depth} & 0.805 & 0.904 & 0.948 & 0.144 & 0.560 & 0.068 \\ 
PrDepth~\cite{prdepth} & 0.800 & 0.926 & 0.969 & 0.151 & 0.460 & 0.063 \\ 
Ours & \textbf{0.853} & \textbf{0.941} & \textbf{0.972} & \textbf{0.123} & \textbf{0.436} & \textbf{0.051} \\ \hline
\end{tabular}
    \caption{Quantitative evaluation on the ZJU-L5 dataset. Our method outperforms all baselines for monocular depth estimation, depth completion, and depth super-resolution.}
    \label{tab:main}
\end{table}

\subsection{Comparison with State-of-the-Art}
Since we are the first to utilize L5 signals and color images to predict depth,
there is no existing method for a direct comparison.
Therefore, we pick three types of existing methods and let them make use of information from L5 as fully as possible. The first method is monocular depth estimation, where we use the depth information of L5 to align the predicted depth globally. The second method is depth completion, where we assume that each zone's mean depth lies at the zone's centroid to construct a sparse depth map as the input. The third method is depth super-resolution, where we consider the L5 signals as an $8\times8$ low-resolution depth map, with each pixel (zone) corresponding to a region of the image. Since the state-of-the-art RGB-D method is sensitive to the sparsity of the input points, we re-trained these methods for a fair comparison.

Table~\ref{tab:main} summarizes the comparison between ours and these three types of methods. For all metrics, our method achieves the best performance among all methods, which indicates that our network customized with respect to the underlying physics of the L5 is effective in learning from the depth distribution.

\begin{figure}[!t]
\begin{center}
    \includegraphics[width=1.0\linewidth]{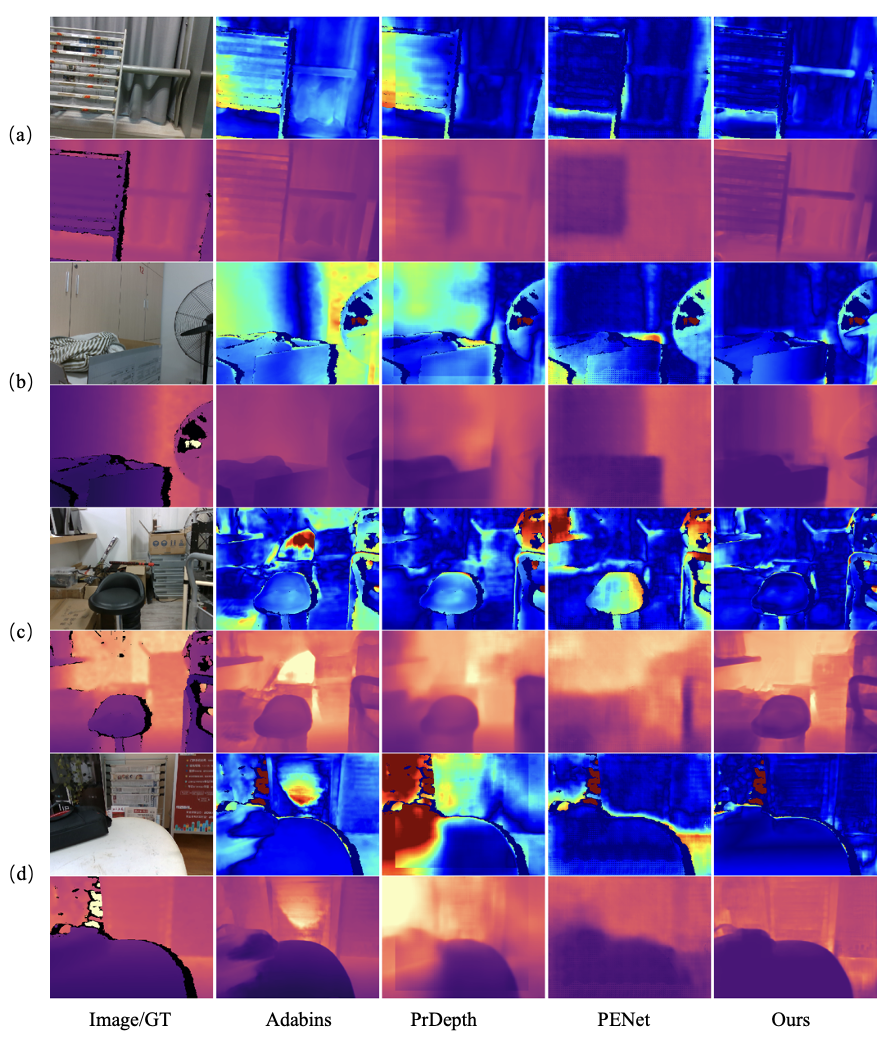}
\end{center}
\caption{Qualitative comparison on ZJU-L5 dataset with error map. Monocular estimation method \cite{adabins} tends to make mistakes 
on some misleading textures.
Guided depth super-resolution \cite{prdepth} and completion \cite{penet} produce overly blurry depths that are lack of geometry details.
In contrast, our method learns to leverage the high resolution color image and low quality L5 reading, and produces the most accurate depths with sharp object boundaries.} 
\label{fig:qualitative-error}
\end{figure}

Fig.~\ref{fig:qualitative-error} shows the qualitative comparison of our method with other solutions for our task~\cite{adabins,prdepth,penet}.
Overall, our method produces the most accurate depth as reflected by the error map.
The monocular estimation method \cite{adabins} can produce sharp object boundaries, however tends to make mistakes for regions with ambiguous textures.
Guided depth super-resolution \cite{prdepth} and completion \cite{penet} by design are easier to maintain plausible depth measurement, but the output depths are often overly blurry and lack geometry details.
In contrast, our method learns to leverage the high-resolution color image and low-quality L5 readings, producing the most accurate depths that are rich of details.

  
\begin{table}[!t]
    \centering

    \tabcolsep 10pt

\begin{tabular}{lrrr}
\hline
Models & \multicolumn{1}{l}{$\delta_1\uparrow$} & \multicolumn{1}{l}{REL$\downarrow$} & \multicolumn{1}{l}{RMSE$\downarrow$} \\ \hline
Mean-Var PointNet & 0.434 & 0.298 & 0.669 \\
Five-channel Input & 0.619 & 0.251 & 0.583 \\
Feature Concat & 0.825 & 0.140 & 0.454 \\
w/o Patch-Dist-Corr & 0.749 & 0.182 & 0.512 \\
w/o Img-Self-Attn & 0.835 & 0.133 & 0.456 \\ 
w/o Img-Dist-Attn & 0.840 & 0.135 & 0.446 \\ 
Uniform Sampling & 0.849 & 0.127 & 0.439 \\
w/o Refine & 0.850 & 0.126 & 0.462 \\ 
\hline
Full & \textbf{0.853} & \textbf{0.123} & \textbf{0.436} \\ \hline
\end{tabular}

    \caption{Ablation studies. We evaluate our method with each design or network component turned off. Overall, our full model achieves the best performance, which indicates the positive contribution from all design choices.}
    \label{tab:ablation}
\end{table}

\subsection{Ablation Studies} \label{sec:ablation}

To understand the impact of each model component on the final performance, we conduct a comprehensive ablation study by disabling each component respectively. The quantitative results are shown in Table~\ref{tab:ablation}. There is a reasonable drop in performance with each component disabled, while the full model works the best.

\noindent\textbf{Learning Directly from the Mean/Variance.} 
We implement two baseline methods which learn directly from the mean and variance of the depth distribution.
For the first one, we change the input to a five-dimensional tensor, which consists of RGB, mean depth and depth variance, named ``Five-channel Input''.
For the second one, we extract features directly from the mean and variance instead of sampled depth, named ``Mean-Var PointNet''.
The performance of these two baselines drops significantly compared to our full model, which indicates the effectiveness of our distribution feature extractor and the fusion module.

\noindent\textbf{Compared with direct feature concatenation.} 
We also replace our Transformer-based fusion module with direct feature concatenation (but retain our proposed feature extractor).
It shows that our fusion module performs better than the direct concatenation,
which benefits from the fact that our strategy can better gather features from totally different modalities and boost the overall accuracy by propagating features in a global receptive field.

\noindent\textbf{Cross-attention without Considering Patch-distribution Correspondence.} 
We re-train a model by relaxing the constrain on cross-attention, name ``w/o Patch-Dist-Corr''.
Specifically, we conduct cross-attention between all distribution features and image features without considering patch-distribution correspondence.
The performance degradation shows the importance of considering this correspondence.

\noindent\textbf{Impact of multiple attention mechanisms.} We train models without image self-attention (``w/o Img-Self-Attn'') or image-to-distribution attention (``w/o Img-Dist-Attn'') respectively that are proposed in Section.~\ref{sec:fusion-module}.
The performance drop indicates that the attention modules positively contribute to our fusion model.

\noindent\textbf{Impact of probability-driven sampling.} We compare our methods trained with uniform sampling and probability-related sampling. The experiment indicates it brings an improvement of 0.3cm in terms of RMSE by considering the distribution probability.
We report the impact of sampling points' number in the supplementary.

\noindent\textbf{Impact of Refinement Module} We also study the impact of the refinement module by replacing it with a simpler decoder consisting of two convolutional layers that output bin-widths vector and linear coefficient respectively. 
The overall performance drops but not much, which indicates, though the refiner helps, the majority improvements are brought by our distribution feature extractor and the fusion network.

\begin{figure}[!t]
\begin{center}
    \includegraphics[width=1.0\linewidth]{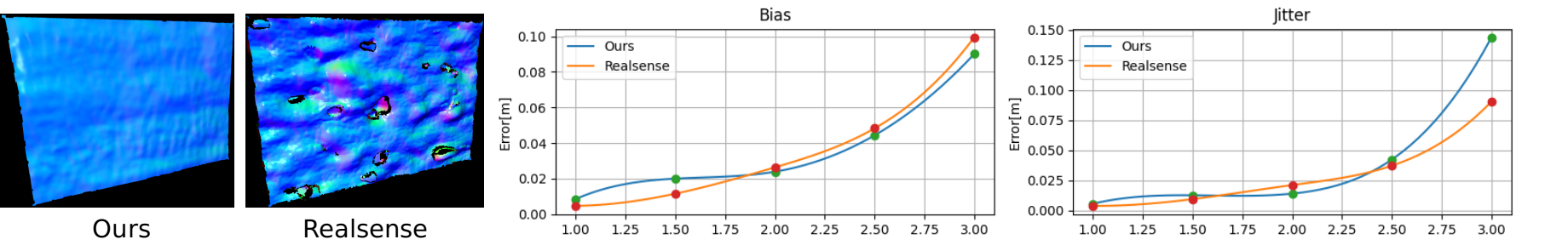}
\end{center}
\caption{Quantitative comparison with RealSense. Our method improves the raw L5 reading to a quality on par with commodity-level RGB-D sensor as reflected by the bias and jitter.} 
\label{fig:comp-with-RealSense}
\end{figure}

\subsection{Quantitative comparison with RealSense}
In this section, we compare our methods with RealSense quantitatively using traditional metrics in the area of stereo matching~\cite{active-stereo-net}, such as jitter and bias. Specifically, we recorded multiple frames with the device in front of a flat wall at distances ranging from 1000 mm to 3000 mm. In this case, we evaluate by comparing to ``ground truth'' obtained with robust plane fitting.
We compute bias as the average $\text{L}1$ error between the predicted depth and the ground truth plane to characterize the precision and compute the jitter as the standard deviation of the depth error to characterize the noise. Fig.~\ref{fig:comp-with-RealSense} shows the comparison between our method and RealSense, together with visualizations of point clouds colored by surface normals. 
It can be seen that at a close range (less than three meters), our method achieves a similar and even better performance than RealSense. But as it approaches the upper range limit of L5, the jitter of our method increases dramatically.
Overall, it indicates that our method improves the raw depth readings of the L5 to a quality (both resolution and accuracy) on par with a commodity-level depth sensor (\ie, the Intel RealSense D435i) in the working range of L5.

\section{Conclusion and Future Work}
In this work, we show that it is feasible to estimate high-quality depth, on par with commodity-level RGB-D sensors, using a color image and low-quality depth from a light-weight ToF depth sensor.
The task is non-trivial due to the extremely low resolution and specific measurements of depth distribution, thus requiring a customized model to effectively extract features from depth distribution and fuse them with RGB image.
One limitation of our method is that it is not fast enough for real-time performance.
It is promising to further optimize the network complexity such that the system can run without much extra cost of energy consumption, or to further extend the system for more applications such as 3D reconstruction or SLAM.

\noindent\textbf{Acknowledgment.} This work was partially supported by Zhejiang Lab (2021PE 0AC01) and NSF of China~(No. 62102356).

\clearpage

\bibliographystyle{splncs04}
\bibliography{egbib}
\end{document}